\documentclass[10pt,twocolumn,letterpaper]{article}

\usepackage[pagenumbers]{cvpr}   

\usepackage{microtype}

\definecolor{cvprblue}{rgb}{0.21,0.49,0.74}
\usepackage[pagebackref,breaklinks,colorlinks,allcolors=cvprblue]{hyperref}

\def\paperID{20}
\def\confName{CVPR}
\def\confYear{2026}

\title{Automated Report-Derived Oncology VQA Benchmark for Evaluating Vision-Language Models on 3D Medical Imaging\thanks{Code and benchmark: \url{https://github.com/Bol901/Med_VQA_Benchmark_Skill}}}

\author{%
Bo Liu$^{1,2,3}$ \quad Hanxue Gu$^{2}$ \quad Xiangru Li$^{1,2}$ \quad Zheren Zhu$^{1,2}$ \quad Jacob Ellison$^{1,2}$\\
Kang Wang$^{2}$ \quad Janine M. Lupo$^{2}$ \quad Yang Yang$^{2}$ \quad Hui Lin$^{3}$\\[0.6em]
$^{1}$UCSF--UC Berkeley Joint Graduate Program in Bioengineering\\
$^{2}$Department of Radiology, UCSF \quad $^{3}$Department of Radiation Oncology, UCSF
}

\begin{document}
\maketitle
\begin{abstract}
Evaluating vision-language models (VLMs) on medical images requires benchmarks that are clinically grounded, scalable, and controlled for evaluation confounds. Existing public benchmarks are limited in scale, manually annotated, or potentially leaked into VLM pretraining corpora. We present an automated agent-driven pipeline that generates multiple-choice VQA datasets directly from paired private radiology reports and 3D oncology imaging, producing two complementary question types: RADS-style questions deterministically derived from clinician-defined reporting schemas, and radiology report-derived questions generated by an LLM from radiologist findings and verified against the source report. Applied to four in-house cancer cohorts, the pipeline yields an instance-contamination-controlled benchmark without per-question human annotation. Zero-shot evaluation of six VLMs reveals no dominant model and substantial headroom across all cells. A blind ablation reveals that visual reliance is highly dataset-specific: liver Report-derived questions genuinely require the image, while Lung CT is essentially solvable without it—the leading closed model exceeds its sighted accuracy on Lung CT when blinded—indicating that even private clinical data does not guarantee a contamination-controlled read of visual capability. The pipeline is released as an open agent skill for in-house redeployment.
\end{abstract}

\section{Introduction}
\label{sec:intro}

Vision-language models (VLMs) have advanced rapidly on general visual tasks, but their reliability in clinical medical imaging remains poorly understood. Existing medical VQA benchmarks face two persistent confounds: limited scale when manually annotated~\cite{vqarad,pathvqa}, and contamination risk when web-sourced, with MIRAGE~\cite{mirage} showed that VLMs answer public medical VQA with minimal accuracy drop even when images are removed. Two properties of clinical oncology imaging further raise the bar. First, the imaging is volumetric: 2D-input VLMs must answer questions from a single representative slice of a 3D acquisition that radiologists read across slices and sequences. Second, every case is a confirmed cancer patient, so questions concern lesion characterization, disease extent, and treatment-relevant features, not presence vs. absence of obvious abnormality, raising the visual-reasoning bar relative to standard medical VQA built on heterogeneous case mixes.\newline
We make three contributions: (1) an automated pipeline that generates dual-path multiple-choice VQA from paired private reports and images, combining schema-driven RADS-style questions with LLM-generated, report-grounded questions; (2) an instance-contamination-controlled benchmark spanning four in-house 3D oncology cohorts; and (3) a zero-shot evaluation of six VLMs that reveals dataset-specific failure modes and a sharp split between visual and language-prior contributions to apparent performance.
\section{Related Work}
\label{sec:related}

\paragraph{Medical VQA benchmarks.}
Early datasets such as VQA-RAD~\cite{vqarad}, SLAKE~\cite{slake}, and PathVQA~\cite{pathvqa} relied on manual annotation and are limited in scale, while larger sets like OmniMedVQA~\cite{omnimedvqa} and PMC-VQA~\cite{pmcvqa} draw on web and literature sources whose public release raises contamination concerns for modern VLMs.\newline
\textbf{Vision-language models in medicine.}
MedPaLM-M~\cite{medpalmm} and LLaVA-Med~\cite{llavamed} adapted general VLMs for medical tasks, and MedGemma~\cite{medgemma} represents a purpose-built medical VLM trained on clinical data; we evaluate this family alongside frontier general-purpose VLMs in a strictly zero-shot setting.\newline
\textbf{Benchmark contamination.}
MIRAGE~\cite{mirage} demonstrated that VLMs can answer public medical VQA questions with minimal performance drop when images are removed, implicating language priors and data leakage. Our use of private clinical reports paired with single-institution imaging controls for instance-level leakage; as we show in §4.3, distribution-level pretraining familiarity remains a confound.

\section{Method}
\label{sec:method}

\begin{figure*}[!htbp]
  \centering
  \includegraphics[width=0.59\linewidth]{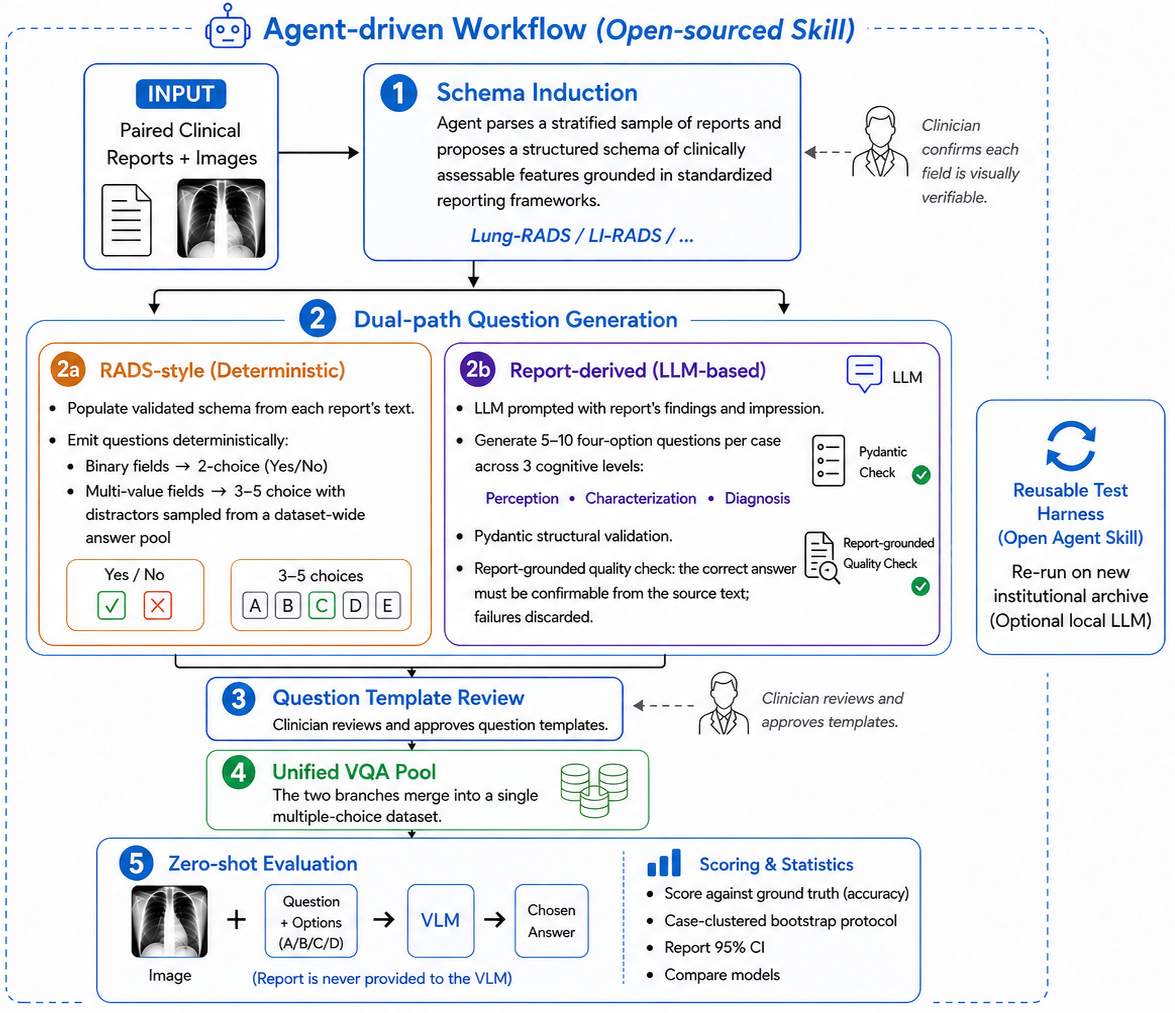}
  \caption{Agent-driven pipeline for constructing a contamination-free radiology VQA benchmark. A dual-path generation strategy produces clinically grounded multiple-choice questions from paired reports and images, followed by clinician validation and zero-shot evaluation. The workflow is released as a reusable test harness for institutional re-deployment.}
  \label{fig:pipeline}
\end{figure*}

\Cref{fig:pipeline} summarizes the end-to-end workflow; the remainder of this section details each stage.

\subsection{Dataset}
We apply our pipeline to four in-house oncology cohorts (brain MRI, liver MRI, liver CT, lung CT); every case is a confirmed cancer patient. Brain MRI is a hybrid configuration: imaging derives from the publicly available PDGM cohort~\cite{ucsfpdgm} paired with private institutional reports. The other three cohorts are fully private from the same institution. Case and question counts appear in \cref{tab:main}. All data use operates under an approved IRB protocol; closed-weights models were accessed via privacy-preserving institutional API endpoints and open-weights models on-premise, with no clinical text leaving the institution.

\subsection{RADS-style Questions}
For each dataset, an LLM induces a schema of clinically relevant features from a stratified sample of reports, grounded in established frameworks (LI-RADS~\cite{lirads} for liver, Lung-RADS~\cite{lungrads} for lung, and an institutional RANO-aligned~\cite{rano} schema for brain). Each field corresponds to a visual finding directly assessable from the image, and clinicians validate the schema for visual answerability before use. An LLM then populates the fields from each report (binary → yes/no; multi-value → 3–5 choices with dataset-wide distractors). Anchoring extraction to a pre-validated schema keeps the pipeline clinically valid, controllable, and reproducible.

\subsection{Report-derived Questions}
Report-derived questions are generated by prompting an LLM with the parsed findings and impression sections to produce multiple-choice questions spanning the radiologist's clinical content. Each question undergoes structural validation and a report-grounded check confirming that the correct answer is explicitly supported by the source text; failures are discarded. The pipeline runs automatically with human review at only two checkpoints: extraction schema confirmation and question template approval. The report-grounded check filters factually unsupported answers but cannot fully eliminate residual LLM authoring bias, which we treat as an inherent limitation of automated question generation.

\subsection{Evaluation Protocol}
\label{sec:method:eval}
Accuracy is the fraction of valid model responses matching ground truth; invalid items are excluded. 95\% confidence intervals are based on case-clustered bootstrap ($B{=}2000$, seed 42), which resamples cases rather than questions to reflect generalization to new patients; we report mean $\pm$ half-width throughout. Because question pools mix 2- to 5-option items, per-cell chance varies; we report a random baseline computed as the mean of 1/k across each pool's choice-count distribution.
\section{Experiments}
\label{sec:experiments}

\begin{table*}[!t]
  \centering
  \footnotesize
  \setlength{\tabcolsep}{4pt}
  \renewcommand{\arraystretch}{0.95}
  \caption{Zero-shot accuracy by dataset and question type. Top rows give per-dataset case counts and RADS/Report-derived question counts. Values are mean $\pm$ half-width of the 95\% case-clustered bootstrap CI ($B{=}2000$, seed 42). Best per column \textbf{bold}.}
  \label{tab:main}
  \begin{tabular}{lcccc}
    \toprule
                          & Liver CT                   & Liver MRI                  & Lung CT                    & Brain MRI                  \\
    \midrule
    Cases                 & $1000$                     & $531$                      & $500$                      & $480$                      \\
    Questions (RAD/LLM)   & $5978/8901$                & $4209/5221$                & $3085/3773$                & $4174/7656$                \\
    \midrule
    \multicolumn{5}{l}{\textit{(a) RADS-style questions}} \\
    \midrule
    Random baseline       & $0.328$                    & $0.289$                    & $0.460$                    & $0.465$                    \\
    Claude Opus 4.6       & $0.453 \pm 0.028$          & $\mathbf{0.571 \pm 0.020}$  & $0.541 \pm 0.028$          & $\mathbf{0.693 \pm 0.018}$  \\
    GPT-5                 & $0.277 \pm 0.029$          & $0.259 \pm 0.020$          & $0.498 \pm 0.017$          & $0.607 \pm 0.022$          \\
    GPT-5.2               & $0.405 \pm 0.024$          & $0.410 \pm 0.021$          & $0.511 \pm 0.022$          & $0.658 \pm 0.020$          \\
    MedGemma1.0-27B          & $\mathbf{0.486 \pm 0.010}$  & $0.336 \pm 0.017$          & $0.524 \pm 0.015$          & $0.654 \pm 0.015$          \\
    MedGemma1.5-4B           & $0.425 \pm 0.017$          & $0.512 \pm 0.018$          & $0.511 \pm 0.017$          & $0.556 \pm 0.014$          \\
    Qwen3-VL-30B          & $0.481 \pm 0.012$          & $0.296 \pm 0.015$          & $\mathbf{0.589 \pm 0.016}$  & $0.622 \pm 0.013$          \\
    \midrule
    \multicolumn{5}{l}{\textit{(b) Report-derived questions}} \\
    \midrule
    Random baseline       & $0.315$                    & $0.324$                    & $0.313$                    & $0.250$                    \\
    Claude Opus 4.6       & $0.617 \pm 0.020$          & $\mathbf{0.587 \pm 0.019}$  & $\mathbf{0.697 \pm 0.030}$  & $\mathbf{0.681 \pm 0.018}$  \\
    GPT-5                 & $0.406 \pm 0.022$          & $0.415 \pm 0.018$          & $0.662 \pm 0.017$          & $0.661 \pm 0.020$          \\
    GPT-5.2               & $0.440 \pm 0.019$          & $0.386 \pm 0.019$          & $0.612 \pm 0.033$          & $0.635 \pm 0.024$          \\
    MedGemma1.0-27B          & $\mathbf{0.645 \pm 0.012}$  & $0.348 \pm 0.014$          & $0.473 \pm 0.016$          & $0.613 \pm 0.012$          \\
    MedGemma1.5-4B           & $0.478 \pm 0.015$          & $0.549 \pm 0.015$          & $0.493 \pm 0.016$          & $0.523 \pm 0.013$          \\
    Qwen3-VL-30B          & $0.440 \pm 0.011$          & $0.278 \pm 0.015$          & $0.563 \pm 0.014$          & $0.600 \pm 0.011$          \\
    \bottomrule
  \end{tabular}
\end{table*}

\subsection{Setup}
We evaluate six VLMs zero-shot: three closed frontier models (Claude Opus 4.6, GPT-5/5.2), one open-weights frontier model (Qwen3-VL-30B), two medical-specialist models at different scales (MedGemma1.0-27B, MedGemma1.5-4B). Models receive the medical image plus question and options; they never see the radiology report. Accuracy is computed in \cref{sec:method:eval}.

\subsection{Main Results}
\Cref{tab:main} reports zero-shot accuracy across all four datasets and both question types, with a per-cell random baseline. No single model dominates: medical-pretrained MedGemma1.0-27B wins both Liver CT cells, Claude Opus 4.6 takes the cells that demand interpretive integration (Liver MRI, Brain MRI, Lung Report-derived), and Qwen3-VL-30B is the only model to top Lung RADS-style. \textbf{Liver MRI RADS-style is near-chance for half the model set}: GPT-5, Qwen3-VL-30B, and MedGemma1.0-27B sit within statistical reach of random baseline, indicating that contrast-rich multi-sequence MRI is not where current 2D-VLM pretraining transfers cleanly.

\subsection{Blind Ablation: Disentangling Visual from Language-Prior Contributions}
\label{sec:blind}
We probe whether models rely on visual evidence by re-running one closed model (GPT-5) and the strongest specialist (MedGemma1.0-27B) without any image input, holding question text, options, and prompt constant with a fixed subsample of N=2000 questions per cell. \Cref{tab:blind} reports blind accuracy and the sighted-minus-blind drop $\Delta$. The deltas vary by an order of magnitude across cells for the same model, exposing that visual reliance is highly dataset-specific. Three regimes emerge.

\textbf{Visual evidence dominates on liver Report-derived items}. Both models lose 0.19-0.30 accuracy when blinded on Liver CT and Liver MRI Report-derived questions, confirming that liver lesion characterization genuinely requires image input.

\textbf{Lung CT is essentially solvable without the image}. GPT-5's blind accuracy exceeds its sighted accuracy on both Lung CT question types. We attribute this to substantial pretraining familiarity with public chest-CT corpora, which provide strong question-text-to-answer associations independent of any individual image.

\textbf{Brain MRI sits between}. Blind accuracy remains high but generally below that of sighted, consistent with strong pretraining priors for categorical brain-tumor labels, likely amplified by the imaging deriving from the publicly available brain tumor PDGM cohort, yet the image still contributes a measurable signal.

One caveat: LLM-generated distractors may carry textual cues that lift blind accuracy above chance even where the image is essential. The cell-by-cell variation in $\Delta$, however, points to dataset-level pretraining familiarity and inherent visual reliance as the primary drivers, with distractor effects roughly constant across cells. The two-model, fixed-subsample design gives a directional signal of contamination rather than tight per-cell estimates; a sweep across all six models on the full question pool would refine them.

\begin{table}[!htbp]
  \centering
  \footnotesize
  \setlength{\tabcolsep}{3pt}
  \renewcommand{\arraystretch}{0.95}
  \caption{Blind ablation. Models receive question + options with the image replaced by a blank input. $\Delta = \text{sighted}-\text{blind}$ uses sighted point estimates from \cref{tab:main}; small or negative $\Delta$ indicates language-prior solvability, large $\Delta$ indicates the image is essential.}
  \label{tab:blind}
  \begin{tabular}{llcccc}
    \toprule
                & & \multicolumn{2}{c}{GPT-5} & \multicolumn{2}{c}{MedGemma1.0-27B} \\
    \cmidrule(lr){3-4} \cmidrule(lr){5-6}
    Dataset     & Q.\ type         & Blind   & $\Delta$  & Blind   & $\Delta$  \\
    \midrule
    Liver CT    & RADS-style       & $0.228$ & $+0.049$  & $0.380$ & $+0.107$  \\
    Liver CT    & Report-derived   & $0.159$ & $+0.247$  & $0.350$ & $+0.295$  \\
    Liver MRI   & RADS-style       & $0.183$ & $+0.077$  & $0.379$ & $-0.044$  \\
    Liver MRI   & Report-derived   & $0.229$ & $+0.186$  & $0.346$ & $+0.002$  \\
    Lung CT     & RADS-style       & $0.520$ & $-0.022$  & $0.455$ & $+0.069$  \\
    Lung CT     & Report-derived   & $0.688$ & $-0.026$  & $0.440$ & $+0.034$  \\
    Brain MRI   & RADS-style       & $0.540$ & $+0.067$  & $0.502$ & $+0.152$  \\
    Brain MRI   & Report-derived   & $0.615$ & $+0.046$  & $0.394$ & $+0.219$  \\
    \bottomrule
  \end{tabular}
\end{table}
\begin{figure*}[!t]
  \centering
  \includegraphics[width=0.8\linewidth]{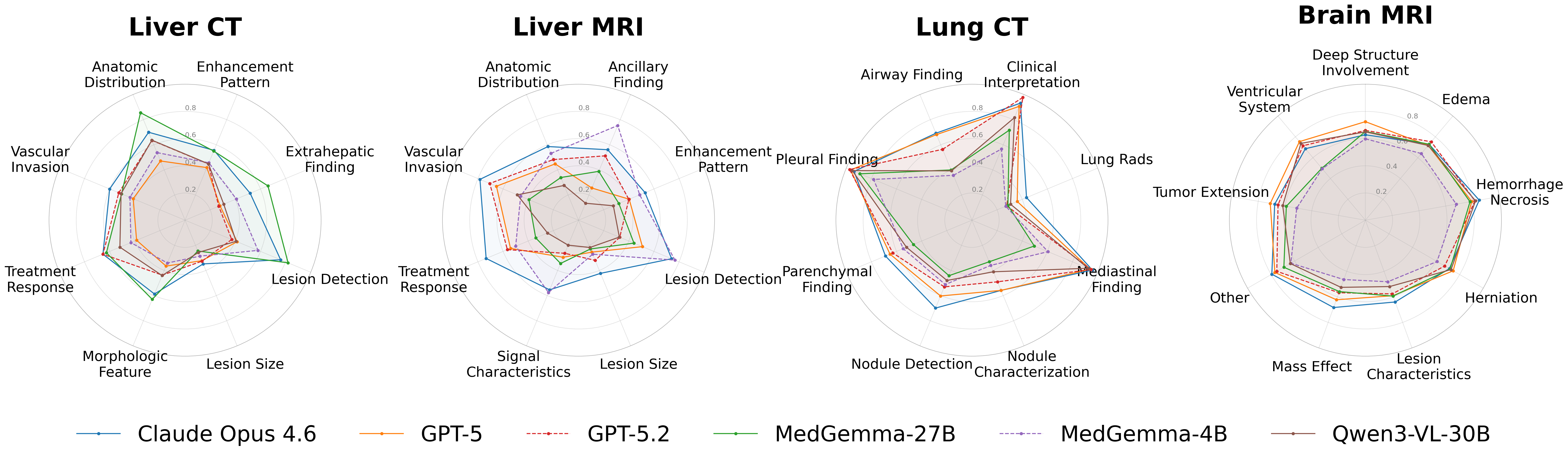}
  \caption{Per-subcategory accuracy on Report-derived questions across the four datasets and six VLMs. Each panel is a polar plot whose axes are the dataset's clinical subcategories; longer radii are better.}
  \label{fig:spider_subcat}
\end{figure*}
\section{Analysis}
\label{sec:analysis}

\subsection{RADS-style vs.\ Report-derived Questions}
The relationship between RADS-style and Report-derived accuracy in \cref{tab:main} is model-dependent rather than uniformly ``RADS is easier.'' Frontier general-purpose models (Claude Opus 4.6, GPT-5, GPT-5.2) score consistently higher on Report-derived items, specialists (MedGemma-27B/4B) are roughly balanced across the four datasets, and Qwen3-VL-30B is the only model that consistently favors RADS-style. RADS-style items on Liver CT and Lung often demand fine-grained visual discrimination, whereas Report-derived items key off textual entities and qualifiers that frontier models can partially solve through verbal-reasoning shortcuts, which is consistent with the blind ablation on Lung CT, where GPT-5 retains its full Report-derived accuracy when blinded. Notably, GPT-5.2 does not uniformly outperform GPT-5 across cells, suggesting recent upgrading within the OpenAI family does not transfer cleanly to private 3D oncology VQA.

\subsection{Cross-dataset Patterns}
The four datasets behave very differently, traceable to imaging physics and 2D-input constraints. \textbf{Brain MRI is uniformly easiest}, attributable partly to multiparametric T1-post-contrast and FLAIR sequences providing strong contrast against a stereotyped intracranial background, and partly to imaging deriving from the public PDGM cohort (paired reports remain private), where the questions themselves are not contaminated, but the imaging distribution may be familiar. \textbf{Liver MRI is hardest}, because clinical reading integrates multiple sequences that 2D VLMs cannot consume, and most models drop to or below random on Liver MRI RADS-style. \textbf{Liver CT is hard for a different reason}: lesion-to-parenchyma contrast is intrinsically low, so even single-slice information demands fine-grained visual discrimination—confirmed by the blind ablation. \textbf{Lung CT performance reflects familiarity with the training data rather than visual reasoning} (§4.3). MedGemma1.0-27B is strongest on Liver CT but the weakest specialist on every MRI cell except brain, suggesting a CT-leaning visual prior.

\subsection{Subcategory-level Performance}
\label{sec:subcat}
\Cref{fig:spider_subcat} breaks down Report-derived accuracy by clinical subcategory across the four datasets. Three patterns emerge.

\textbf{(i)~Easy and hard subcategories are largely shared across models.} Lesion-presence judgments (Lesion Detection on liver, Mediastinal/Pleural on Lung, Hemorrhage on Brain) sit near the top across models. Hard subcategories are also stable: Lesion Size on liver, Lung-RADS scoring, and Mass Effect / Lesion Characteristics on Brain MRI are uniformly the weakest cells.

\textbf{(ii)~The split is between qualitative recognition and quantitative or graded discrimination.} Recognition tasks (`is X here?') are within reach; numerical or ordinal judgments (lesion size, Lung-RADS category, mass-effect grading) are not, even when the relevant region is visible. Lung-RADS is the clearest example: despite being the most clinically structured Lung subcategory, every model scores below $0.45$ on it, suggesting that learned discrete-scoring rubrics with fine size/morphology thresholds remain out of reach.

\subsection{The Pipeline as a Portable Harness}
\label{sec:harness}
The generation skill is released as an open agent workflow. Any institution can re-run it on paired reports and images—optionally with a locally hosted LLM to keep clinical text on-premises—to build a modality-specific in-house benchmark. 
\section{Conclusion}
\label{sec:conclusion}
We presented an automated pipeline for generating clinically grounded VQA from paired private report-image archives, applied it to four 3D oncology cohorts, and zero-shot evaluated six VLMs. No model approaches saturation, where the highest cell sits near 0.70, and visual reliance varies sharply by dataset: liver lesion characterization genuinely requires the image, while Lung CT is essentially solvable without it, with the leading closed model exceeding its sighted accuracy when blinded. Even private clinical data does not guarantee a contamination-controlled read of capability. The pipeline is released as an open agent skill for institutional redeployment.

{
    \small
    \bibliographystyle{ieeenat_fullname}
    \bibliography{main}
}

\end{document}